# Well Tops Guided Prediction of Reservoir Properties using Modular Neural Network Concept: A Case Study from Western Onshore, India

Soumi Chaki[1], Akhilesh K. Verma[2], Aurobinda Routray[1], William K. Mohanty[2*],

Mamata Jenamani[3]

[1]Department of Electrical Engineering, Indian Institute of Technology Kharagpur

[2]Department of Geology and Geophysics, Indian Institute of Technology Kharagpur

[3]Department of Industrial and Systems Engineering, Indian Institute of Technology Kharagpur

[*]Email address of corresponding author: wkmohanty@gg.iitkgp.ernet.in

## ABSTRACT

This paper proposes a complete framework consisting pre-processing, modeling, and post-processing stages to carry out well tops guided prediction of a reservoir property (sand fraction) from three seismic attributes (seismic impedance, instantaneous amplitude, and instantaneous frequency) using the concept of modular artificial neural network (MANN). The dataset used in this study comprising three seismic attributes and well log data from eight wells, is acquired from a western onshore hydrocarbon field of India. Firstly, the acquired dataset is integrated and normalized. Then, well log analysis and segmentation of the total depth range into three different units (zones) separated by well tops are carried out. Secondly, three different networks are trained corresponding to three different zones using combined dataset of seven wells and then trained networks are validated using the remaining test well. The target property of the test well is predicted using three different tuned networks corresponding to three zones; and then the estimated values obtained from three different networks are concatenated to represent the predicted log along the complete depth range of the testing well. The application of multiple simpler networks instead of a single one improves the prediction accuracy in terms of performance evaluators– correlation coefficient, root mean square error, absolute error mean and program execution time. Then, volumetric prediction of reservoir properties is carried out using calibrated network parameters. This stage is followed by post-processing to improve visualization. Thus, a complete framework, which includes pre-processing, model building and validation, volumetric prediction, and post-processing, is designed for successful mapping between seismic attributes and a reservoir characteristic. The proposed framework outperformed a single artificial neural network in terms of reduced prediction error, program execution time and improved correlation coefficient as a result of application of the MANN concept.

1. **Introduction**

Modeling of petrophysical characteristics from well logs and seismic data plays a crucial role in petroleum exploration. Two major challenges are faced while interpreting and integrating different kinds of datasets (mainly, well logs and seismic data); 1) nonlinear and diverse natures of reservoir variables associated with the subsurface systems, and 2) absence of any direct relationship between seismic and well log signals from a theoretical perspective. Similarly, calibration of a functional relationship between a reservoir characteristic and predictor seismic attributes is an intricate task. Linear multiple regression and neural networks are popular among statistical techniques for reservoir modeling from well logs and seismic attributes (Nikravesh and Aminzadeh, 2001; Bosch et al., 2005). Lately, several computation intensive artificial intelligence (AI) methods such as artificial neural network (ANN), neuro-fuzzy, self-organizing map (SOM), committee machine and learning vector quantization (LVQ) have attained recognition as the potential tools to solve nonlinear and complex problems in the domain of reservoir characterization (Fung et al., 1997; Nikravesh et al., 1998; Nikravesh and Aminzadeh, 2001; Nikravesh and Hassibi, 2003; Hamada and Elshafei, 2010; Bosch et al., 2010, Karimpouli et al., 2010; Majdi et al, 2010; Asadisaghandi and Tahmasebi, 2011; Tahmasebi and Hezarkhani, 2012). Permeability, porosity, fluid saturation and, sand and shale fractions are some of the fundamental reservoir characteristics spatially distributed non-uniformly. Despite the difference in theory and computation, most of the modelling algorithms are applied for the same purpose. On the contrary, a single technique can serve as a potential tool for solving different problems. For instance, artificial neural network (ANN) has been applied in several areas of science and technology for different objectives. Moreover, different categories of neural network architectures are capable to solve nonlinear problems, however, as complexity of the problem increases due to enhancement in the number of inputs or the complex nature of the predictor variables, the performance of the network decreases rapidly.

Several researchers have claimed that application of modular networks and hybrid networks show better performance compared to a single algorithm (Smaoui and Garrouch, 1997; Auda and Kamel, 1999; Jiang, et., al., 2003; Al-Bulushi et al., 2009; Asadi et al., 2013; Al-Dousari et al., 2013). In particular, due to heterogeneous nature of subsurface system (or reservoir variables), application of a single network for complete depth range of a well may not be sufficient to achieve adequate prediction accuracy. In this context, module based networks, so-called modular artificial neural networks (MANN), are well suited to solve complex nonlinear problems. Moreover, the concept of modularity is applied in many fields to divide a complex problem into a set of relatively easier sub problems; then, the smaller sub-problems are solved by modules; finally, the obtained results are combined to achieve the solution of the main problem (Jiang et al., 2003; Mendoza et al., 2009; Ding et al., 2014; Sanchez and Melin, 2014). The module-wise division is carried out based on different clusters and classes of the dataset. This modularity concept is implemented individually or along with another machine learning algorithms to solve different types of problem. Lithological information extraction from integrated dataset is a major step in the reservoir modeling. Additionally, it is essential to investigate how 3D seismic information is associated to production, lithology, geology, and well log data. It is envisaged that the incorporation of 3D seismic data along with well logs can provide better insights while extrapolating reservoir properties away from the existing wells (Doyen, 1998; Bosch et al., 2010).

MANN is previously used in reservoir characterization. Fung et al., (1997) used to predict petrophysical properties from a set of well log data. The smaller networks are constructed corresponding to different classes obtained from trained LVQ network. However, seismic attributes are not considered in the study by Fung et al., (1997). A similar work has been carried to predict permeability from multiple well logs such as spectral gamma ray, electrical resistivity, water saturation, total porosity etc. recorded from four closely spaced boreholes using modular neural network (Tahmasebi and Hezarkhani, 2012). The dataset is divided into three sets– 70%, 15%, and 15% for training, validation, and testing respectively. Despite improvement in the prediction results compared to a single network, this study suffers some inherent limitations. Firstly, seismic attributes are not included in the dataset. Secondly, blind prediction is not carried out. Thirdly, selection of the number of networks and number of hidden layer neurons are not guided by any particular theory. The best network is finalized by trail-and-error framework. In the present study, we have addressed the drawbacks of it (Tahmasebi and Hezarkhani, 2012) and design our proposed framework accordingly.

In oil exploration, different lithological classes, clusters, zones of interest in terms well tops and horizons are identified from the preliminary analysis of well logs and it is integrated with seismic data of the same region. In the past, many classification and clustering techniques have been used to make different classes of data depending on their variability (Fung et al., 1997; Bhatt and Helle, 2002; Kadkhodaie-Ilkhchi et al., 2009; Verma et al., 2012). Recently, the concept of chaotic time series data analysis namely dynamic programming (Lisiecki and Lisiecki, 2002), synchronization methods (Donner and Donges, 2012, Verma et al., 2014) are applied to assess similarity between the pattern of the well log data, and which lead towards the identification of similar zones among the wells. Generally, zonation of the logs is carried out manually by experienced geoscientists. Above nonlinear approaches are aimed to provide information of similar patches in the log data or similar zones in different wells; and hence have potential application in the reservoir characterization. In the present study, well tops are identified from combination of well logs and accordingly different zones are marked on the log data. In the literature, it has been claimed that modular (multi-nets) systems have the advantage of being easier to understand or modify as per requirement. Geoscientists' guided zone-wise division of a well log can assist in target evaluation after training several models using zone wise divided training patterns yielding improved prediction accuracy (Fung et al., 1997; Bhatt and Helle, 2002). Hence, present study is focused on module (multi-nets) wise prediction of the reservoir property (i.e. sand fraction) for better accuracy.

We can encapsulate the work done in this paper as a motivated outcome of the concepts of modularity and synchronization together. Firstly, the idea of well tops guided division of the dataset is evolved from synchronization or similarity. The similarity between well logs sections belong to a certain horizon is more compared to that of the similarity between multiple complete length logs. Next, the modularity concept enables to divide a complex problem into a set of relatively simple sub-problems. The borehole data are available at specific well locations; whereas seismic attributes are acquired over the area of interest. If a functional relationship can be established between seismic and well log signals (petrophysical properties), then, the variation of the reservoir characteristic over the area can be predicted from the seismic attributes itself. As the predictor and target signals are from two different domains, therefore, a single artificial neural network structure may not be able to successfully obtain the mapping function between these two types of signals, which is the current research problem in this domain (Tahmasebi and Hezarkhani, 2012). In this paper, we have attempted to devise a complete framework, with the objective of overcoming the limitations of previous studies.

This paper demonstrates the application of MANN concept to predict sand fraction from seismic attributes (seismic impedance, instantaneous amplitude, and instantaneous frequency). In this study, two well tops (namely Top1 and Top2) are identified after analysing well logs and seismic data. In the process of mapping sand fraction from seismic attributes, first, seismic attributes are extracted from 3D seismic cube at eight well locations. Next, integration of seismic and borehole data are carried out using time-depth relationship information at the available well locations. The pre-processed master dataset is then divided into three zones based on the two well tops such as $1^{st}$ available patterns to Top 1, Top 1 to Top 2, and Top 2 to last available data pattern. In the model building and validation stage, first, three networks have been designed for three different zones separately. Sand fraction and three seismic attributes correspond to seven wells are used for training and testing, and patterns correspond to the remaining well are used for blind prediction. The satisfactory performance in blind testing encourages to carry out volumetric prediction of sand fraction using the three trained models. Then, results evaluated by three different networks (zone-wise) are merged to form a volumetric cube containing the estimated sand fraction values across the study area along the entire depth range. After model building and validation, a post-processing is carried out to improve the visualization quality.

The rest of this paper is organized as follows. Section 2 presents the detail description of the dataset used in this study. Section 3 demonstrates the methodology based on modular artificial neural network (MANN) concept. Then, the proposed framework combining pre-processing, model building and validation, volumetric prediction, and finally, post-processing stages are presented in Section 4. Section 5 represents analysis of the experimental results. Finally, Section 6 concludes the finding of the present study, its application and future scope.

## 2. Description of dataset

The well logs and seismic data used in this paper are acquired from a hydrocarbon field located at the western onshore of India. The borehole dataset includes basic logs such as gamma ray, resistivity, density and other derived logs such as sand fraction value, permeability, porosity, water saturation, etc. Conversely, the seismic dataset includes different attributes, i.e., seismic impedance, instantaneous frequency, seismic envelope, seismic sweetness, etc. This paper involves seismic impedance, instantaneous amplitude and instantaneous frequency to model sand fraction using an integrated dataset of seismic and sand fraction signals available at eight well locations.

The distribution of seismic impedance over the study area at a particular time instant is demonstrated in Fig. 1. In the same figure, placements of the eight wells used in this study are also identified. It can be seen from Fig. 1 that the boreholes are distributed over the study area.

The lithological properties along the depth range of a well vary in a non-linear and heterogeneous fashion. The variations of lithological properties along depth for well 4 and well 5 are captured in Fig. 2. Two well tops namely Top 1 (red line) and Top 2 (green line) are shown in the figure. Integrated dataset corresponding to seven wells are used for learning of the three zone-wise prediction models and data from the eighth well is used for validation purpose.

## 3. Methodology based on MANN concept

In recent years, artificial neural networks (ANNs) are widely used to solve nonlinear modeling problems in the fields of science and technology such as computer science, electronics, mathematics, geosciences, medicine, physics, etc., (Poulton, 2002). ANN and its hybrid approaches have also been proven to be useful in the nonlinear mapping of reservoir properties from well logs and seismic data (Poulton, 2002; Deshmukh and Ghatol, 2010). ANN performs satisfactorily in non-linear data mapping, pattern recognition and classification problems; however, the execution speed is slow in case of working with a large dataset. Therefore, there is significant scope of improvement in order to accelerate the training process by modifying the basic algorithm while not compromising with the prediction accuracy. Hence, modular artificial neural network (MANN), which is a special category of ANN based on data categorization, is introduced as a potential tool for machine learning with efficient estimation capability and high speed (Anand et al., 1995; Fung et al., 1997; Lu and Ito, 1999; Bhatt and Helle, 2002; Deshmukh and Ghatol, 2010). The concept of modularity is derived from the principle of divide and conquers. Here, a complex computational task is subdivided into smaller and simpler subtasks. Each local computational model performs an explicit, interpretable and relevant job according to the mechanics of the problem involved. Finally, the output of the model will be the combination of individual results of dedicated local computational systems. In this approach, module wise networks are trained and tested, and the outputs of all modules are integrated to achieve complete sequence of the target variable.

Present study discusses an application of MANN concept for prediction of a reservoir property from seismic attributes. Well tops represent abrupt changes in the log data which corresponds to the changes in lithology denoting the corresponding zone boundaries. In this case, two well tops (Top 1 and Top2) are marked on the logs of petrophysical properties by expert geologists which in turn segments a log into three zones: Zone1: starting of log to Top 1, Zone 2: between Top 1 and Top 2, Zone 3: Top 2 to end of the log. Previous studies (e.g., Verma et al., 2014) reported that the similar zones in a well log reveals similar characteristic. Based on this hypothesis, the number of modules are decided as three same as the number of zones. Therefore, the master dataset combining seismic and borehole data is first divided into three zones (Z-1, Z-2 and Z-3) based on well tops guidance. Fig. 3 represents a schematic diagram depicting application of modular artificial neural network (MANN) concept for the present study. Fig. 4 depicts the preparation steps of the zone-wise database starting from data compilation to zone-wise division. Each of the input patterns contain three inputs (seismic impedance, instantaneous amplitude, and instantaneous frequency) pertaining to three input layer nodes for all models along with single output layer neuron denoting target sand fraction.

| **ALGORITHM I: MODULAR NEURAL NETWORK APPROACH** |
|---|
| **Task :** Mapping between target property (sand fraction) and input predictors (Seismic attributes) |
| **Input**: Seismic attributes and sand fraction |
|     a) Integration of seismic attributes and sand fraction signals |
|     b) Zone/module wise division of data (*n*: total number of zones) |
|     c) Decide blind well: k |
|     d) Data partitioning – training and testing set |
|     e) Data normalization based on min-max normalization OR z-score normalization |
|     f) **for** $\underline{i} = 1$ to $n$ (*i*: number of zones) |
|     g) Selection of network structure (number of neuron, training algorithm) |
|     h) Initialization of weights and biases, maximum epoch: $iter_{max}$, min error: $error_{min}$ |
|     i) **for** epoch=1: $iter_{max}$ |
|     j) Modify weights and biases following selected training algorithm |
|     k) Calculate root mean square error |
|     l) **if** root mean square error $\leq error_{min}$ ∥ epoch=$iter_{max}$ break, **else** epoch= epoch+1, **end** |
|     m) **end for** |
|     n) Test the network using testing patterns of well k |
|     o) **if** testing is satisfactory go to step p) **else** go to step g) |
|     p) Freeze the network structure : $MANN_i$ ($i = 1$ to $n$, here $n=3$) |
|     q) Save the network structures and parameters for minimum error |
|     r) **end for;** |
| **Output**: Three sets of calibrated network parameters (weights and biases) w.r.t. three zones |

Dataset from seven wells are used for training of the network, and then trained network is used to blindly model sand fraction for remaining one well. Separate training is carried out for each of the three modular networks keeping the learning algorithm and transfer functions same for all three networks. The optimal model is obtained by minimizing the root mean square error between network output and target using selected state-of-art learning algorithm for each case. The testing of each model is carried out by using the zone-wise divided testing patterns corresponding to eighth well which is not included in the training set. Proposed MANN approach for the present study is described in Algorithm I. Thus, three mapping functions are obtained using MANN approach (Fung et al., 1997), and these are corresponding to three zones (Z-1, Z-2 and Z-3). These three trained networks are further used to obtain predicted sand fraction log for the whole study area. As indicated in Fig. 3, the predicted sand fraction logs from each modular network are concatenated to obtain the complete log profile. The obtained inputs-target relationships are used to estimate the lithological properties over the whole study area from seismic attributes.

## 4. The proposed framework

Seismic data are collected over a large study area, whereas well logs are available at specific well locations in the same region. Furthermore, the vertical resolution of seismic attributes is inferior compared to that of the well logs due to larger sampling interval. In general, the seismic data are helpful to model a reservoir; however, it is difficult to estimate vertical distribution of reservoir properties with the help of seismic signals (Doyen, 1998; Bosch et al., 2010). Therefore, guidance of both - seismic and well logs is necessary to characterize a reservoir property with high resolution in vertical and horizontal directions. For example, sand/shale fraction, porosity, permeability and saturation are important petrophysical properties used in the interpretation of hydrocarbon reserves in details. Therefore, modeling of any such petrophysical characteristic has crucial importance in this research domain.

In this paper, sand fraction is estimated from three seismic signals (seismic amplitude, impedance and instantaneous frequency) using modular artificial neural network (MANN) concept. This paper proposes a framework, which includes pre-processing, modeling and validation, volumetric prediction, and post processing, to carry out sand fraction modeling successfully.

*4.1 Pre-processing*

This step involves integration and normalization of target (sand fraction) and predictor variables (seismic amplitude, impedance and instantaneous frequency), followed by data partition into training and testing set.

A. *Integration of seismic and well log signals*

Integration of signals from different domains with the help of heuristic knowledge from human experts plays a major role in reservoir characterization (Nikravesh, 2004). Therefore, the first task in pre-processing is integration of seismic (which is in time domain) and well log signals (which is in depth domain) at each available well location. First, we extract the seismic attributes at eight available well locations. Then, data points in well log signals carrying missing values are excluded. It is followed by conversion of logs from depth to time domain using suitable velocity profile resulting from well-to-seismic tie. Then, the mismatch in sampling intervals of these two data sources (seismic and well logs) is observed. Specifically, band limited seismic attributes are sampled at an interval of 2 milliseconds, whereas the sampling interval of well logs is ~0.15 milliseconds for this particular dataset. Since, the sampling intervals of both the data are different, we apply Nyquist–Shannon sampling theorem (Shannon, 1949), which states that a band limited signal can be completely reconstructed from the samples, to reconstruct seismic attributes at each time instant corresponding to the well logs using cubic spline interpolation method (Neill, 2002). Due to removal of missing values from logs, the dataset is not uniform anymore. Finally, the dataset uniformly re-sampled at an interval of 0.10 milliseconds.

*B. Data normalization*

Data normalization plays a crucial role for tuning the performance of the machine learning algorithms. During the experimentation with the current dataset and the choice of the algorithm, finally, we proposed different normalization schemes for predictor and target variables. The predictors and target variables are normalized using the Z-score and min-max normalization, respectively.

*C. Data partition*

This is a common approach in machine learning algorithms to divide a dataset into training and testing sets for learning and validation, respectively. In this study, a combined dataset of seismic and well log signals corresponding to seven boreholes is used for training the networks whereas data of the remaining eighth well is used for testing the networks.

In this study, well tops guided zone wise prediction is carried out using the concept of modular artificial neural network (MANN). Fig. 4 depicts a workflow for integration and division of the dataset into three separate zones (Z-1, Z-2, and Z-3) for further modeling of reservoir properties.

*4.2 Model building and validation*

The learning starts after completion of pre-processing of the working dataset. Three networks corresponding to each of the three depth zones (Z-1, Z-2, and Z-3) are trained and tested. Each of the networks has three predictor variables corresponding to presence of three input nodes in network structure and a single output node representing target sand fraction. In this study, we opted for a single hidden layer for all cases. There are two types of transfer functions in the Back Propagation Neural Network (BPNN), one is in the hidden layer and other one is in the output layer. Selection of activation functions and training algorithm plays crucial role in training of the network. In the present study, hyperbolic tangent sigmoid is used in the hidden layer (Kalman et al., 1992; Leshno et al., 1993); however, output layer uses log-sigmoid transfer function. In these types of iterative processes, the connecting weights are updated using the back propagation till the global minimum error is achieved. Conjugate gradient method is an advanced and effective method for error minimization (Haykin, 1999). Here, scaled-conjugate-gradient-back-propagation is selected over several learning algorithms for its speed and simplicity (Haykin, 1999) to train the networks. As mentioned earlier, min-max normalization is carried out to normalize the target variable between 0.2 and 0.8 to avoid saturation region of the log-sigmoid function. Number of neurons in the hidden layer and epochs are initialized with small values and gradually increased keeping the improvement of fitting between target and predicted sand fraction in view. However, the number of hidden layer neurons cannot be indefinitely increased; the possibility of over fitting should be avoided by keeping the maximum number of trainable parameters at least fifteen times lower than the number of available training patterns (Haykin, 1999).

Three separate networks are designed and trained for three depth zones, and finally trained networks are used for blind prediction. The performance of the trained networks are quantified in terms of four performance evaluators namely correlation coefficient (CC), root mean square error (RMSE), absolute error mean (AEM) and program execution time. It is important to carry out statistical analyses of the errors involved in the model. The calibrated networks which performed well in blind prediction in terms of four aforementioned performance evaluators are saved and used in the next step, i.e., volumetric prediction.

*4.3 Volumetric prediction*

This step is essential for visualization of reservoir characteristic at and away from the boreholes after prediction from seismic attributes is carried out. As no direct relationship between seismic and well logs is evident in theory, which might be inherent, it is a challenging task to estimate lithological properties across the study area from seismic signals. Therefore, it would be beneficial for the geoscientists if a mapping between seismic and reservoir properties could be carried out by deriving a relationship between these two types of data from integrated dataset of seismic and lithological parameters at available well locations using MANN concept. The horizon or well tops information of the study area is available. Therefore, the dataset containing predictor attributes throughout the study area are segregated into three parts according to well tops information. Then, for each zone, predicted sand fraction log is generated from seismic signals using tuned network parameters corresponding to the particular zone. Thus, a set of three logs is available for each particular trace point. These three logs can be concatenated according to obtain complete sand fraction log at a particular trace point. Hence, sand fraction logs are predicted for the study area from seismic input using tuned networks. After that, necessary initiatives are adopted to represent the prediction result for whole area. In this paper, visualization at a specific in-line is demonstrated after predicting the sand fraction from three seismic attributes (seismic impedance, instantaneous amplitude, and instantaneous frequency) over the area. In parallel, input attributes are also visualized across the in-line. It can be seen from the figures presented in the results section that predicted sand fraction requires post-processing step to improve the visualization quality.

*4.4 Post Processing*

In this study, the predicted sand fraction is smoothened using moving average filter (Oppenheim and Schafer, 1989). The necessity of the post processing step is established by comparing the variation of seismic attributes and estimated sand fraction.

---

**ALGORITHM II** : MOVING AVERAGE FILTER

---

**Task** : Reduce noise in predicted sand fraction volume

**Input** : Predicted sand volume matrix $X$, window size $w$

   a) Initialize : Window size, $w$;

   $x_{j,k}$ - pixel value at $(j,k)$, $I_{j,k}^w$ be a window of size $w \times w$ centered at $(j,k)$

   b) Compute $i_{j,k}^{mov,w}$ –average of the pixel values in $I_{j,k}^w$

   c) Replace $x_{j,k}$ by $i_{j,k}^{mov,w}$, thus obtain $X_{filt}$

   d) If result is satisfactory, then stop, else go to step a).

**Output** : Filtered sand fraction matrix $X_{filt}$

---

Every matrix element in predicted sand fraction volume is considered as a pixel and smoothened using moving average filter respective to neighborhood of pixels within selected window size following Algorithm II. In specific cases, where some of the neighborhood cell values are missing for a particular element, those missing values are replaced by NaN (not a number). For example, edge of the input matrix is filtered following above procedure. In this paper, the window size of the moving average filter used to smooth the predicted sand fraction is selected as 3×3 empirically.

## 5. Results and discussion

The proposed workflow is implemented on a 64 bit MATLAB platform installed in a Intel(R) core (TM) i5 CPU @3.10 GHz computing system having 8.00 GB RAM. The objective of this work is to model sand fraction from seismic attributes using modular artificial neural network (MANN) concept. First, a combined dataset of seven wells is used to train three different neural networks according to depth wise zones (Z-1, Z-2, and Z-3) (see Figs. 3 and 4), then, the trained networks are validated using the dataset of the remaining well. The three predicted log sections for test well corresponding to each zone (Z-1, Z-2, and Z-3) are merged to obtain complete sand fraction log. As it is a convention to present well log in time domain, hence, predicted sand fraction logs are represented in depth domain. Fig. 5 represents superimposed plots of target and network predicted sand fraction values for Zone 1, 2, and 3, respectively for well 6 only. Close observation on Fig. 5 reveals that the predicted logs follow the target ones with acceptable correlation coefficients (0.8058 for Z-1; 0.7699 for Z-2; and 0.8841 for Z-3). These high values of correlation coefficients indicate good prediction by the proposed framework. Similar results are obtained for other wells also.

Fig. 6 describes the variation of input seismic attributes and predicted sand fraction at an in-line corresponds to well 6. It can be observed from Figs. 6(a)–(c) that the input attributes change smoothly throughout the study area. On the other hand, networks predicted sand fraction variation is not smooth. This uneven variation in predicted sand fraction necessitates inclusion of a post-processing algorithm. We opt for a moving average filter based algorithm with a 3×3 window size. Implementation of the filtering technique on predicted sand fraction reduces noise in it. Comparing Fig. 6(d) with Fig. 6(e), it can be observed that the variation of the latter is smoother than former. Thus, a realistic presentation of sand fraction variation over an area is obtained.

The correlation coefficients obtained by blind testing using three networks corresponds to three zones (Z-1, Z-2, and Z-3), and their average are compared with blind prediction coefficient using a single ANN for the overall depth range. Figs. 7–10 represent the results of performance comparison of the proposed workflow with an ANN in terms of performance evaluators for well 2, 4, and 6. For example, in case of Well 4, first, the dataset is segregated into three sections following the well tops guided zonation. Then, three sets of training patterns are generated combining the samples belong to Well 1, 2, 3, 5, 6,7 and 8 for each of the segregated dataset in the previous stage. Three networks are initialized and trained using the training patterns corresponding to three zones (Z1, Z2, and Z3). Next, the calibrated networks are validated using testing patterns belong to Well 4 for each zone separately. In case of CC, RMSE, and AEM, average performance of the proposed framework is expressed by carrying out mean of the respective measures belong to three individual models. Fig. 7 and Figs. 9–10 demonstrate correlation coefficients, root mean square errors, and absolute error means respectively by the three individual models (Model 1, Model 2, and Model 3 for Z-1, Z-2, and Z3 respectively), their average performance, and the single ANN model used for the overall depth range. In contrary, the total program execution time is resultant of summation of the three individual models. Fig. 8 depicts individual and total program execution times in seconds taken by the three models workflow along with that of the single ANN associated with whole depth range. As smaller networks deal with simpler structures and smaller dataset, the accepted accuracy is achieved within a reduced execution time in case of MANN approach.

## 6. Conclusions

The objective of the present study is to establish well tops guided prediction of a reservoir property using MANN concept over a single network while working on a large and complex dataset. This paper has presented the performance of the proposed workflow along with ANN by blind estimation of sand fraction from seismic attributes (amplitude, instantaneous amplitude, and instantaneous frequency). It is evident from the presented results that the proposed workflow has outperformed ANN in terms of higher correlation coefficient, reduced error measures and low program execution time by successfully calibrating a functional relationship between seismic and well log signals corresponding to each zone. Thus, MANN concept can be selected over ANN in case of complex large dataset. The post-processing on predicted sand fraction improves the visualization realistically.

The contributions of this study are consolidated as follows:
- Fusion of two concepts – similarity between logs belong to same horizon and MANN

- Inclusion of seismic data as predictor variables
- The selection of number of modules based on well-top information
- Blind prediction
- The proposed workflow is established to have better performance within reduced program execution time
- Enhanced visualization by post-processing

Next phase of research may be focused on estimation of 3D geo-cellular model for other characteristics involving seismic and well log signals at available well control points. Here, number of modules are decided based on well tops guided instead of trail-and-error methods. However, the optimum number of hidden layer neurons are not automated till now. Therefore, introduction of evolutionary algorithms along with the proposed framework to automate number of neurons selection can be included in future scope.